# What Students Ask, How a Generative AI Assistant Responds: Exploring Higher Education Students' Dialogues on Learning Analytics Feedback

Yildiz Uzun

UCL Knowledge Lab, University College London, UK, yildiz.uzun.22@ucl.ac.uk

Andrea Gauthier

UCL Knowledge Lab, University College London, UK, andrea.gauthier@ucl.ac.uk

Mutlu Cukurova

UCL Knowledge Lab & UCL Centre for Artificial Intelligence, University College London, UK, m.cukurova@ucl.ac.uk

**Abstract**

Learning analytics dashboards (LADs) aim to support students' regulation of learning by translating complex data into feedback. Yet students, especially those with lower self-regulated learning (SRL) competence, often struggle to engage with and interpret analytics feedback. Conversational generative artificial intelligence (GenAI) assistants have shown potential to scaffold this process through real-time, personalised, dialogue-based support. Further advancing this potential, we explored authentic dialogues between students and GenAI assistant integrated into LAD during a 10-week semester. The analysis focused on questions students with different SRL levels posed, the relevance and quality of the assistant's answers, and how students perceived the assistant's role in their learning. Findings revealed distinct query patterns. While low SRL students sought clarification and reassurance, high SRL students queried technical aspects and requested personalised strategies. The assistant provided clear and reliable explanations but limited in personalisation, handling emotionally charged queries, and integrating multiple data points for tailored responses. Findings further extend that GenAI interventions can be especially valuable for low SRL students, offering scaffolding that supports engagement with feedback and narrows gaps with their higher SRL peers. At the same time, students' reflections underscored the importance of trust, need for greater adaptivity, context-awareness, and technical refinement in future systems.

# 1 INTRODUCTION AND RELATED WORK

The increasing technological developments in LADs in higher education hold the potential to provide data-rich feedback to students on their learning behaviours [28, 32]. By visualising learning process data, LADs could support SRL competencies and help students monitor their progress, reflect on feedback, and make informed decisions about their learning strategies [25, 36, 41]. To frame this role of LADs in relation to learning processes, this study adopted Winne and Hadwin's SRL model that conceptualises learning as a cyclical process of task definition, goal setting, strategy use, and adaptation, guided by metacognitive monitoring and shaped by students' prior knowledge and motivation [36]. Students with higher SRL competence are therefore generally better positioned to use LADs effectively, whereas students with lower SRL competence may be less able to connect feedback to meaningful learning decisions.

Despite the potential of analytics feedback in the learning process, recent research indicated that students often experience challenges in making sense of analytics data, leading to low adoption rates and limited impact on learning outcomes [18, 19]. One key challenge is the disconnection between analytics feedback and the actionable steps students need to take based on it. While LADs present performance metrics, they may lack providing clear strategies to students for translating feedback into behavioural change [2]. Another challenge is that even if feedback is available, students may still struggle with the complexity of visualised data and find it difficult to extract a practical understanding [30]. In practice, these challenges differ in that the former stems from a lack of actionable direction, the latter from the interpretive effort required. Both challenges are closely related to SRL differences, as students with lower SRL competence may find data visualisations overwhelming or fail to derive actionable next steps [25, 33]. These limitations underscore the need for additional scaffolds that can help reduce the gap in engagement with feedback, while leaving open questions about which supports are most effective and whether their benefits extend similarly to students with higher SRL competence. Prior research on feedback literacy also emphasised that students benefit from interactive and iterative discussions to internalise feedback and develop learning strategies [6, 39]. However, such discussions could be resource-intensive, requiring deep expertise in both content and pedagogy, and be difficult to scale when relying solely on human tutors for each student [14].

As a possible solution, implementing dialogue-based interactive features in LADs could help overcome the identified challenges. Since effective engagement with analytics feedback requires students to understand feedback, clarify doubts, reflect on their learning behaviours, and formulate strategies for improvement [35, 39]. Dialogue with GenAI on analytics feedback in LADs has the potential to facilitate deeper sense-making of, and promote more effective engagement with feedback, leading to better learning outcomes.

## 1.1 Integrating GenAI into LAD to Support Engagement with Feedback

In recent years, advances in GenAI have offered many potential solutions for supporting students, particularly through its integration into LADs. Conversational large language models can act as mediators of the engagement gap between students and analytics feedback, providing on-demand, personalised support to facilitate dialogue, engagement, and reduce frustration and cognitive overload associated with complex dashboards [27, 37]. Unlike static dashboards, a GenAI-integrated LAD can respond dynamically to students' queries about feedback, facilitate active engagement while supporting reflection and deeper comprehension through conversation [14, 39]. Among recent innovations, the VizChat, an open-source chatbot developed by Yan et al. [38], has demonstrated how a GenAI chatbot can provide contextually relevant explanations and clarifications of LAD visualisations with multimodal GenAI capabilities. Building on this foundation, recent research conducted by Jin et al. [15] compared two distinct versions of VizChat in a controlled experiment, with higher education students in a single session: a 'conventional' chatbot that provided reactive, context-specific responses to student queries, and a 'scaffolding' chatbot that offered proactive guidance through structured prompts. The results of the study showed that chatbots in both conditions significantly improved students' understanding of complex analytics feedback presented in the LAD.

These findings not only demonstrate the value of GenAI integration in enhancing analytics feedback comprehension but also raise important questions about how students actually utilise such conversational tools beyond controlled settings. Recent empirical studies, for example, have indicated discrepancies between students' initial perceptions of GenAI-implemented feedback tools and their actual use in practice to interpret feedback [13]. While initial responses were generally positive towards these tools, subsequent engagement data during the term frequently revealed modest utilisation rates, suggesting that students' perceived and actual utility of these tools diverged significantly. This advocates the importance of examining students' uptake of GenAI-implemented feedback tools, and the quality and function of GenAI responses in shaping engagement with analytics feedback [40]. Recent work has also argued that GenAI should act as a scaffold to promote students' feedback literacy and self-regulation rather than replace human judgment [7, 13]. This perspective emphasises a need for balanced integration of GenAI tools that complement rather than replace human input [8] and improve students' agency in making informed decisions about their learning [10]. What remains less understood is why exactly and how GenAI support seems particularly valuable for students with lower SRL competence, and whether its contribution for higher SRL students is more limited. To understand the reasons underlying this divergence and to support the development of student agency in feedback engagement processes for both groups of students, it is crucial to analyse authentic interactions between students and GenAI assistants in real-world educational contexts [13]; thus, we could improve these tools to better meet diverse learner needs.

One way to pursue this deeper investigation is through qualitative analyses of conversational data, which could provide a powerful lens for understanding the cognitive and metacognitive processes students employ when interacting with the GenAI assistant about their analytics feedback [20]. Detailed conversation analysis can shed light on how students frame their questions, articulate their difficulties, and utilise GenAI assistance to enhance their understanding of analytics feedback while also revealing how these questions differ across levels of SRL competence. Such analysis is also critical for refining GenAI-based systems by examining the quality of responses and aligning their functionalities more closely with student needs. As Dai et al. [9] argue, assessing not just whether feedback is delivered, but how it is interpreted and applied, directly informs the educational value and usability of these tools in authentic settings [37].

Given these critical gaps, our research focuses explicitly on conversation analysis within an authentic educational environment. It explores not only *how* students utilise these AI tools in the context of their learning analytics feedback but also the *quality* of the GenAI assistant's responses in addressing student questions and the *perceptions* of students regarding this novel form of support, with the following research questions: **RQ1.** To what extent do the types of questions students ask the GenAI assistant about their learning analytics feedback vary based on their level of SRL competence? **RQ2.** To what extent does the GenAI assistant effectively address the range of questions posed by students? and **RQ3.** How do students perceive the impact and usability of a GenAI assistant within the LAD, and what improvements do they suggest?

## 2 METHODOLOGY

### 2.1 Educational Context and Participants

This study took place in a campus-based postgraduate module on Educational Technology, involving thirty-four students (8 male, 26 female) from diverse academic backgrounds, such as computer science, education, and design sciences. Over 10 weeks, the course focused on topics related to educational technology design. Participants ranged in age from under 22 years ($n = 2$) to over 30 years ($n = 11$), with the majority aged between 22–25 years ($n = 13$) and 26–30 years ($n = 8$). All students shared similar experiences with the course's technological resources, such as the learning management system (Moodle) and documentation platform (Google Docs) and possessed comparable proficiency levels in statistical analysis and programming. Institutional ethics approval was obtained before data collection, and participation was voluntary. At the start of the course, students were fully informed about the study and provided written consent to take part.

Personalised analytics feedback was generated using students' interaction data from Moodle and Google Docs. The processed trace data were visualised using line graphs, bar charts, network diagrams, and aggregated interaction counts to support the interpretation of students' learning activity. Then, this feedback was shared with students twice during the semester, in weeks 4 and 7, via a custom-built, web-based LAD developed specifically for the module [33]. To further scaffold engagement with the analytics feedback, the application incorporated a GenAI assistant capable of interacting with students in natural language and providing conversational support and guidance.

### 2.2 The GenAI Assistant Development Process

The GenAI assistant was designed to enhance student engagement with the LAD to initiate real-time, personalised support. This integration aimed to help students better understand and act on the feedback they receive, while improving their SRL competencies. Serving as an interactive tool within the LAD, the GenAI assistant enabled students to ask questions and receive immediate, context-aware responses to help them interpret complex feedback visualisations and suggest strategies for improvement of their SRL behaviours.

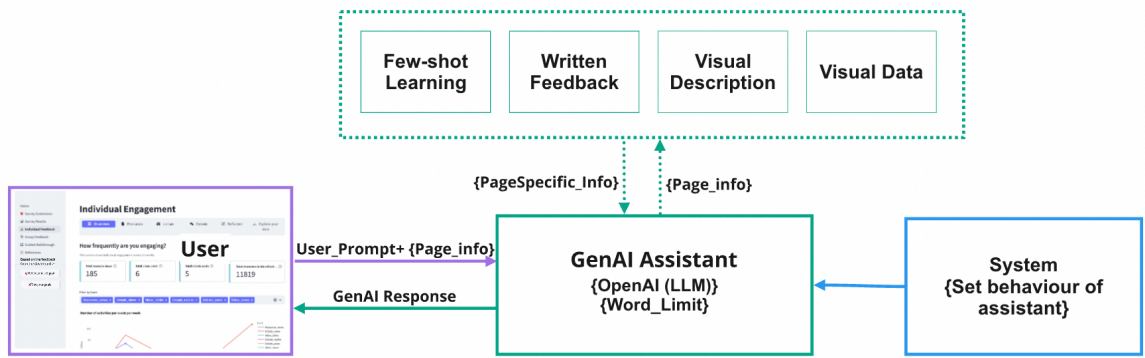

Fig. 1. The GenAI Assistant system architecture.

The GenAI assistant was developed using the OpenAI API (gpt-4, max_tokens = 200, temperature = 1, top_p = 1, openai.com), with specific customisations to ensure it is knowledgeable about the educational context and the specific feedback provided by the LAD. The assistant's responses were configured with a max_tokens parameter and limited each response to 150 words or fewer with a line of prompt for concise and focused answers to maintain students' attention. The temperature value was set to 1 for a balanced variability in the responses to feel both dynamic and natural while staying on-topic. This set of configurations helped ensure that responses were sufficiently comprehensive yet concise, reducing cognitive load for students and helping them retain interest without overwhelming them with excessive information.

When a student interacted with the assistant in a specific part of the LAD, the system generated an input for GenAI assistant that included several components: general system instruction to guide the assistant's behaviour, few-shot learning examples to shape its responses, details about the visuals, written feedback presented on the screen, and relevant student data used in visuals (Fig. 1.). With this combined final prompt consisting of several points, the GenAI assistant was expected to respond students' queries with personalised explanations and suggestions based on their data. The details of the system architecture components are as follows (see Appendix for details[1]).

**The system prompt:** It refers to the initial input text used in the OpenAI API to guide the model's responses. A well-crafted, specific system prompt is a key step, as it directly affects the relevance and accuracy of the model's output. The following system prompt was created for the LAD to maintain a concise context and define the assistant's role: "*As a chat-based assistant, your role is to offer real-time, context-aware responses to students' questions about specific visuals…*"

**Few-shot Learning:** The assistant's responses were adjusted through a few-shot learning approach, which means providing the model with a series of example questions and answers to guide its output. Few-shot examples focused on typical questions that students might ask about their analytics feedback, for example, "*What does this feedback mean for my learning progress?*" The assistant has been primed to respond with guidance that encourages students to reflect on their learning behaviours and consider potential improvements.

**Visual Description:** The predefined descriptions of visual elements in the dashboard (e.g., graph legends, colour codes, and visual cues) were provided to the system to prevent the assistant's misunderstandings when students asked specific questions about the specific visuals without generating misleading or fabricated information.

**Visual Data:** The assistant had access to relevant data points used to create visuals from the student's interactions with the course activities, sourced from platforms like Moodle and Google Docs. It was aimed to personalise responses by including metrics such as completion percentages or frequencies.

**Written Feedback:** Incorporating rule-based written feedback into the prompt provided a baseline for interpreting student interaction metrics. This feedback was crafted by specific engagement thresholds and already presented to students on the dashboard. Including it in the prompt informed the assistant's suggestions, preventing it from advising unnecessary additional activity if engagement was already sufficient.

The GenAI assistant was developed through an iterative refinement process, with the researchers testing and adjusting its responses extensively. To protect students' data privacy, the data sent to the GenAI assistant was anonymised, and the settings regarding training the model with user prompts were turned off on the OpenAI API website.

---

[1] https://tinyurl.com/LAK26Appendix

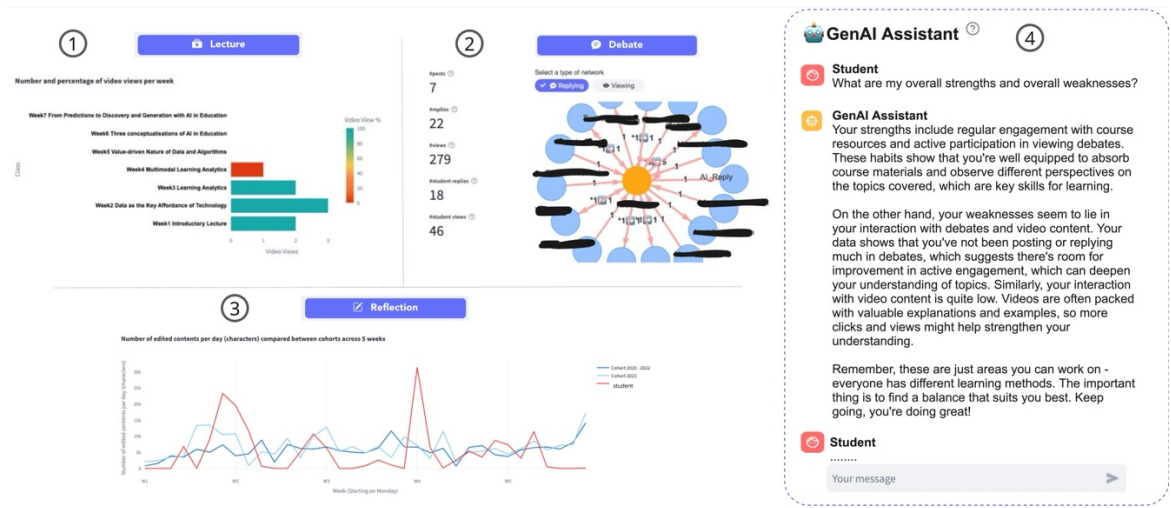

Fig. 2. LAD visualisation examples: 1) a bar chart, representing engagement with prerecorded lecture video; 2) a debate network for online discussion participation; 3) a line graph for reflection writing interaction; and 4) a sample conversation with the GenAI assistant.

## 2.3 Data Collection and Analysis

This study addressed the three research questions through a combination of conversational log data analysis and student reflections. All students' interactions with the GenAI assistant in the LAD were recorded over the semester, a total of 168 queries and accompanying responses (Fig. 2.). Each exchange was time-stamped and linked to the student's session to determine how many students used the assistant, how frequently they engaged, and the nature of their queries. The aim was to capture common interaction patterns, points of confusion in interpreting feedback, and recurrent help-seeking behaviours. During the semester, 22 of the 34 enrolled students engaged with the assistant; therefore, this study focused on them. Two students communicated in their native language (Chinese), and their interactions were translated into English using ChatGPT. The translations were verified by two native speakers to ensure accuracy and contextual fidelity.

**RQ1.** After data cleaning and preprocessing, an inductive thematic analysis was conducted to identify recurring categories and subcategories of students' queries to the GenAI assistant [5]. To capture the multi-faceted nature of some queries, multiple codes were assigned to a single question [4]. To measure SRL competence, students completed a validated SRL questionnaire [31, 33] at the start of the semester. Scores were standardised, and students were grouped into high (N=12) and low (N=10) SRL clusters using k-means clustering, optimised through silhouette analysis to ensure clear separation between groups. These clusters provided the basis for comparing interaction patterns in subsequent analyses. Then, Epistemic Network Analysis (ENA) was used to examine how students with different levels of SRL competencies interacted with the GenAI assistant about their analytics feedback based on identified query types [29]. ENA models the structure of connections between coded dialogue moves rather than their isolated frequency. Using a moving stanza size of three, ENA captured co-occurrences of query types within a sliding conversational window, showing how queries and actions were linked across time [3]. The resulting network mapped these connections as nodes and weighted edges to directly compare between students with high and low SRL competencies.

**RQ2.** To evaluate the GenAI assistant's effectiveness in addressing student queries, each response in the chat log dataset was assessed using the eight criteria in Table 1. In this study, the assistant's responses were conceptualised as a form of feedback, and the evaluation criteria were therefore grounded on established feedback and SRL frameworks [11, 36], alongside previous response evaluation studies [9]. In feedback theory, Hattie and Timperley [11] define effective feedback as clarifying goals (feed up), progress (feed back), and next steps (feed forward). Combined with the SRL model, feedback can be seen as both a cognitive and metacognitive process that supports monitoring, evaluation, and strategy adjustment. To rate each criterion, a 5-point Likert scale (1 = Poor, 5 = Excellent, or NA if not applicable) was used, and initial scores were discussed and refined by the researchers to improve the evaluation criteria. This scoring approach allowed for fine-grained evaluation of multiple dimensions of response quality. For example, a response might score highly in Clarity and Personalisation but lower in Actionability if it explained the analytics feedback well but failed to offer concrete improvement strategies.

Table 1. Criteria list and descriptions to evaluate the GenAI assistant responses.

| Criteria | Description | Theory |
|---|---|---|
| Clarity | Is the response clear, avoiding jargon, and easy for the student to understand? | Clear responses minimise the cognitive load for Task Definition [36], also refer to Readability criteria in [9] |
| Personalisation | Is the response personalised based on the student's context and specific data points? | Goal Setting and Planning phase in SRL [36]. Tailored feedback improves its effectiveness [11] |
| Actionability | Does the response suggest actionable strategies for improvement? | Feed Up and Forward [11], actionable advice aligns with Strategy enactment [36] and ensures improvement. |
| Context Awareness | How effectively does the assistant integrate multiple data points specified in the GenAI Assistant system architecture? Does the assistant understand the data and its implications for the students' learning? | Process-level Feedback [11] involves the integration of all the data points. |
| Motivation and Engagement | Does the response use a supportive tone and encourage students to engage? | Motivational feedback supports engagement and emotional regulation [11]. |
| Reliability | Does the response align accurately with the student's performance and avoid misleading information? | Reliability is essential to trust in feedback. Accurate Feedback Polarity [11] reflects performance and maintains trust. |
| Multi-page Cohesiveness | Do the responses remain cohesive across the pages of the dashboard? | Consistent responses across multiple pages/visuals support students in monitoring and evaluating their learning progress accurately [36]. |
| Error Handling | Does the assistant effectively handle incomplete, conflicting, missing data or out-of-scope questions (not related to analytics feedback)? | When encountering incomplete or conflicting questions, the GenAI assistant should provide clear, specific, and complete answers to help students stay on task, provide feedback, and avoid confusion [36]. |

**RQ3.** To capture students' subjective experiences, students were asked to write around 500 words of individual reflection in week 8 after interacting with the GenAI assistant. The reflection prompts encouraged students to discuss the assistant's potential and value, perceived impact on their learning behaviours, and suggestions for improving its effectiveness and usability. These narratives were analysed using a hybrid approach of inductive and deductive thematic analysis [4, 5], combining data-driven coding with concepts informed by the research focus. The analysis centred on attitudes toward the GenAI assistant, perceived benefits and limitations, and specific recommendations for improvement.

For all the analysis of the queries and reflections, researchers coded a subset of data and calculated inter-rater reliability using Cohen's Cappa, which achieved a greater than threshold of .6 [26]. Disagreements were resolved through discussion, leading to refinements in the codebook. This process ensured consistency and reliability in coding across the full dataset.

## 3 RESULTS

### 3.1 RQ1- Students' Queries

To address this research question, we thematically analysed the content of student queries directed at the GenAI assistant to identify the range and nature of questions posed in relation to their analytics feedback. This analysis revealed seven distinct themes, reflecting the varied ways students engaged with the assistant (see Appendix for details).

**Clarification** (Understanding Learning Analytics Feedback). Students in 17.7% of the queries were trying to understand the "what" and "how" of the analytics feedback and data used to prepare the feedback. Under this theme, many students had sought clarification on some **Definitions,** asking for a term, "*What are assessment views?*", *"How did you analyse debate views?"* and on **Interpretation** of feedback, asking for a summary or the main takeaway, "w*hat are my overall strengths and overall weaknesses?*" These suggested that students have tried to make sense of the feedback by requesting detailed information on how analytics feedback and visualisations were generated and interpreted.

**Personalised Insight.** In 34.5% of the queries, students requested specific suggestions and future steps based on their individual learning data, highlighting the need for actionable analytics feedback. While some questions were **Prescriptive,** asking for a specific, direct solution "*Give me suggestion to improve...*" or "*How can I improve...?*", in some cases, students asked very detailed **Exploratory** questions for strategies, like "W*ill you devise a plan for me to catch up to the tasks that need to be completed over the next 10 days*" then "*Can you tailor this to specific points from my data?*" as a follow-up question. Such interactions demonstrated students' strong desire for specific, data-driven feedback rather than generic recommendations from the assistant about analytics feedback.

**More Content Information.** 16.4% of students' queries focused on a learning activity, or assessment content (e.g., reflective writing, course reading, or discussion content) that was not directly measured or visualised on the dashboard. The content-related queries aimed at seeking help to improve reflective writing and debate discussion, structure them, and understand the assessment criteria. Many students asked for guidance on reflective writing with queries such as "*How can I improve quality of my reflection?*", sometimes with a copy of the content of their writings. Others inquired about assessment, such as "*How will this reflection be assessed?*" This suggested that students sought help for more structured guidance on the content of their artefacts within the LAD.

**Justification.** In some queries (7.3%), students tended to explain their behaviours to the GenAI assistant to justify why they received feedback in a certain way. For example, when a student was not engaging with the reflection writing activity and accordingly feedback indicated low engagement, the student explained the underlying reason and provided evidence to this behaviour by saying *"I did not submit my reflection writing document to the tutor yet... because I need to prioritise the core reading before the class."* Also, students explained their learning preferences and schedules to rationale their circumstances, *"but I want to balance work and rest, so I choose to relax during reading week..."* when the GenAI assistant explained the feedback with data points or recommended certain study routines. These justification-type queries were usually accompanied by other types of queries, such as seeking personalised guidance based on their needs or clarification of the misinterpreted parts in the analytics feedback, which exemplified the active dialogue between students and the GenAI assistant to negotiate and co-construct the meaning of the feedback to share the same understanding of it.

**Emotional Expression.** Students (5.9%) used the assistant not only for guidance but also to express emotion, frustration, and gratitude about their analytics feedback. One student opened a conversation with "*I struggle with retaining some information mentioned during the debates...*", while another shared a preference for reading over videos and concern about receiving low engagement scores. Such exchanges demonstrated how students engaged with the assistant to express themselves, clarify personal learning preferences, and interpret feedback in ways that align with their individual learning approaches. Brief affirmations like "*great, awesome, thanks*" also indicated moments where the assistant met expectations.

**Technical Aspects.** Some queries (8.6%) centred on the technical functioning of the GenAI assistant, the accuracy of the data, and its interpretation within the feedback process. Students questioned how their learning behaviour data was captured and evaluated, for instance, by asking, "*How accurate are the measurements if I open the materials once and download them?*" Others inquired about the assistant's capabilities, posing questions such as "*What are your functions?*" and "*What questions can I ask you?*" These interactions revealed students' engagement with the underlying mechanisms of the feedback tool and highlighted the importance of transparency and clarity for students to be able to use analytics feedback in an informed way.

**Others.** This theme consisted of queries (9.5%) that were not directly related to the analytics feedback but served to manage social interaction with the assistant (e.g., greetings such as "*Hi*" or sign-offs like "*Bye*"). It also encompassed queries that fall outside the intended scope of feedback, including questions about other modules or general questions. While some students also asked about topics within the current module (e.g., "*Can you explain logistic regression?*"), these were still categorised as "Other" if they did not explicitly reference or respond to the analytics feedback.

Based on these seven themes, an **ENA** was conducted to investigate differences in the types of queries students with high and low SRL competence. The Mann-Whitney U test indicated a significant difference in ENA positions between high and low SRL students along the primary dimension (MR1-X axis: 12.0%), which captured the main contrast in discourse networks, $U = 97.00$, $p = .02$, $r = -.62$, with high SRL students (*Mdn* = 0.06, *n* = 12) positioned higher than low SRL students (*Mdn* = -0.17, *n* = 10) (Fig. 3.). This means that the low SRL students appeared more focused on asking for more information about content, seeking clarification, and expressing uncertainty or emotional reactions, possibly signalling a reliance on the assistant for basic understanding or guidance. On the other hand, the high SRL students' queries involved issuing commands, questioning technical aspects, and asking personalised insights tailored to their learning behaviour, suggesting more strategic and autonomous engagement with the GenAI assistant. No significant difference was found along the secondary dimension (SVD2-Y axis: 30.3%), $U = 67.00$, $p = .67$, $r = -.12$, suggesting that meaningful distinctions in discourse networks between SRL groups emerged mainly along MR1.

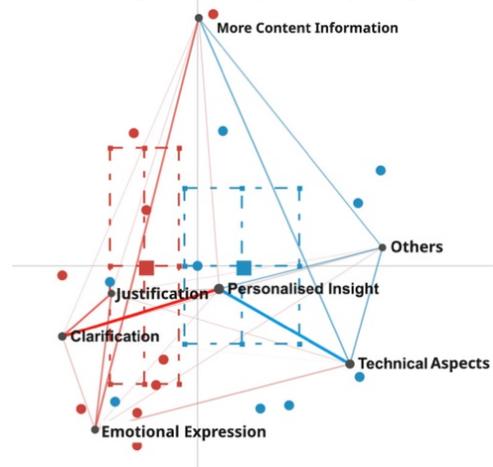

Fig. 3. Comparison graph of query types between low SRL students (**red**) and high SRL students (**blue**).

## 3.2 RQ2- The GenAI Assistant's Response

To address RQ2, we examined how well the assistant's responses aligned with students' informational and reflective needs according to a theoretically grounded framework comprising eight criteria (Table 1). Across all queries, the GenAI assistant's responses were rated highest on Clarity ($M$=4.63, $SD$=0.58), with most responses rated as easy to understand and free of unnecessary jargon. Motivation and engagement ($M$=4.23, $SD$=0.77) and Multi-page Cohesiveness ($M$=4.20, $SD$=0.84) also received high ratings, indicating a generally supportive tone and consistent feedback across interactions. Reliability ($M$=4.12, $SD$=0.89) was rated positively, reflecting alignment between the assistant's feedback and the student's performance data. Moderate ratings were given for Context Awareness ($M$=3.84, $SD$=1.16) and Actionability ($M$=3.72, $SD$=0.81), suggesting variability in how well the assistant integrated multiple data points or offered concrete strategies for improvement. Personalisation criterion ($M$=3.56, $SD$=1.24) was lower, indicating limited tailoring of responses to individual learning contexts. Error handling received the lowest rating ($M$=3.29, $SD$=1.19), showing weaknesses in managing incomplete, conflicting, or other type of questions.

### 3.2.1 The GenAI Assistant Response Quality by Student Query Type

We further investigated the GenAI assistant's response quality by **query type** to identify its strengths and weaknesses in supporting different kinds of student queries. Analysis of ratings, as shown in Fig. 4, revealed consistent strengths in Clarity scores, remained high across all, ranging from $M$=4.47 for "More Content Information" and "Others" queries to $M$=4.81 for "Justification" queries. Personalisation scores varied more widely, from $M$=4.00 in "Others" queries to a low of $M$=2.43 in "Technical Aspects" queries. Actionability was generally moderate, with means ranging between $M$=3.60 and $M$=3.83 across queries. Context awareness varied, with the highest ratings in "Others" ($M$=4.17) and lowest in "Technical Aspects" ($M$=3.27) and "Emotional Expression" ($M$=3.38) queries. Motivation and engagement scores were highest in "Emotional Expression" queries ($M$=4.42) and "Justification" queries ($M$=4.38). Reliability ranged from $M$=3.86 ("More Content Information") to $M$=4.29 ("Personalised Insight"). Multi-page cohesiveness was higher for "Clarification" ($M$=4.47) and "Technical Aspects" ($M$=4.43) queries. Error handling was higher in "Technical Aspects" ($M$=4.25) and "Justification" ($M$=4.00) queries, and lower in "Personalised Insight" ($M$=2.86) and "Emotional Expression" ($M$=3.00) queries.

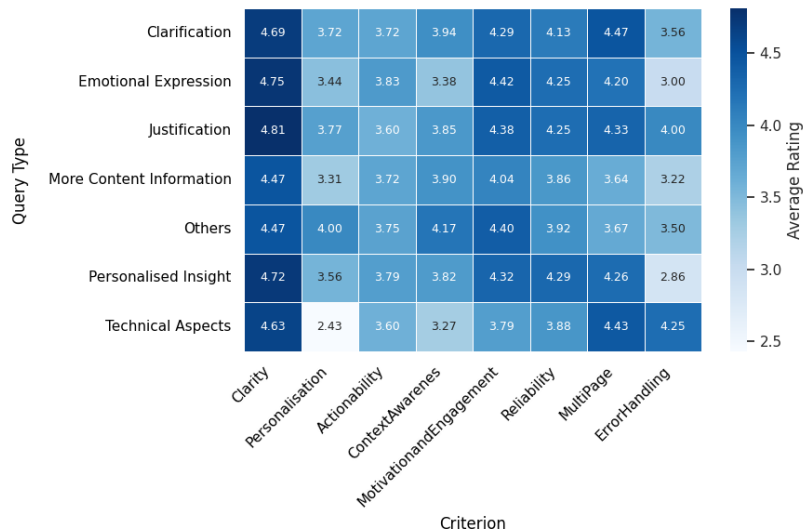

Fig. 4. The GenAI assistant response evaluation by students' query types.

### 3.3 RQ3-Students' Perceptions and Suggestions

The reflections of students on their experiences with the GenAI assistant were analysed under three themes: Affordances, Challenges and Improvements. The findings elicited mixed but meaningful reflections from students, showing their potential to enhance the learning process while underscoring critical areas of improvement (see Appendix for details).

**Affordances.** Students mostly praised the assistant's ability to deliver personalised and context-aware responses derived from their unique data patterns, such as recommending weekly reflection routines, debate participation (e.g., *"The AI recommended that I read more resources to deepen my understanding of the concepts, which would enable me to make more active contributions in discussions"*) or explaining trends in their feedback visualisations (e.g., *"AI explained the nature of my graphs, helping me reflect on factors influencing their shapes"*). They found this personalisation helpful to reflect on learning behaviours and adopt new strategies. Also, the assistant's positive and encouraging tone and ability to highlight strengths based on data points (e.g., *"AI acknowledging my consistent lecture engagement"*) were appreciated for boosting motivation and feeling more confident. This positive attitude resulted in changing study habits, structured study routines, as one wrote, *"I used suggestions to be more proactive about my learning"* and led to a sense of accomplishment.

**Challenges.** Despite the stated benefits, students identified several challenges and limitations that impacted their experience with the GenAI assistant. The first identified challenge was the specificity of responses. Some students found the responses to be "*too broad*" or "*contradictory*", leaving them confused. For example, one student stated that the response was "*helpful but not specific enough*", while another criticised "*inconsistent responses to similar queries*" that diminished the reliability of the assistant. While it was effectively maintaining a supportive environment, some students noted that it was "*overly safe*", which made it difficult to receive constructive criticism. This tendency toward safe responses reduced the value of deeper self-reflections and improvements, especially for students who are already familiar with their data and analytics feedback (e.g., *"I understood the data, so I didn't ask questions"*). This suggested a need for more adaptive systems, shaped by students' level of understanding and offering more advanced insights rather than reiterating existing interpretations. Also, some of the students expressed frustration with technical challenges such as slow response times and its inability to keep past interactions. These issues disrupted the continuity of conversations and resulted in difficulty for students to build on previous responses to similar queries and track their progress over time. Additionally, inconsistent responses to similar questions caused some students to question the assistant's reliability, as noted by a student, *"I encountered some discrepancies between the data provided by the AI system and my actual experiences."*

**Improvements.** However, students saw these challenges as opportunities and offered concrete suggestions to create a more personalised, interactive, and contextually sensitive feedback assistant tool. First, they called for more personalised feedback from the assistant that reflected their backgrounds, prior learning, and specific writing tasks, as one student put it, "*more context-aware feedback… linked to our demographic, personal backgrounds, and reflective writing tasks [content].*" They also stressed the value of a more interactive and proactive dialogue with the assistant, noting that "*the chatbot gave factual responses, not questions, that would support reflection on analytics feedback*" and suggesting that initiating metacognitive conversations could help them "*articulate weekly/long-term objectives*". It was believed that such a conversation would motivate them to engage with the feedback and help them act by setting goals and monitoring them. Therefore, it called for tighter integration of the contents of course activities, proposing to "*integrate individual reflections directly into the platform*" and use GenAI to "*revise learning materials, not just explain feedback,*" for holistic interaction with the assistant. Finally, usability improvements were seen as essential to accomplish all these proposed enhancements. The emergent ones were to "*improve response speed"* and to "*add a history/navigation bar for deep learning based on previous dialogue,*" for smoother and more engaging interactions with the assistant. Overall, these suggested improvements would help transform the assistant from a basic feedback provider to an active learning partner that supports students' reflection, goal-setting, and productive engagement throughout the learning process.

## 4 DISCUSSION

This study explored the conversations taking place in the LAD between students and the GenAI assistant to gain detailed insights into the interaction process. The analysis focused on what students asked the assistant, how it responded, and where it succeeded or fell short in helping them interpret and act on their analytics feedback.

For **RQ1**, the results revealed that students' queries ranged from clarification and recommendation-seeking to self-explanation and emotional expression, highlighting how students with varying levels of SRL competencies interact with feedback when conversational support is available. This moved analytics feedback beyond the traditional assumption of passive monitoring [34] to a dynamic, conversational process. The GenAI assistant was introduced to help students interpret analytics feedback, positioning it as a resource for clarifying visualisations and metrics. The dominance of personalised insight queries suggested that students were not passive consumers of information but actively interpreting their data and planning changes to their learning behaviour with the recommendations, consistent with SRL theory [36, 41]. However, this pattern varied notably by students' level of SRL competence. While low SRL students tended to ask for additional feedback on content, seek clarification, and express uncertainty or emotional reactions, signalling a reliance on the assistant for basic understanding and reassurance, high SRL students, in contrast, issued directive queries, interrogated technical aspects, and requested personalised insights tailored to their data, demonstrating a more autonomous and strategic use of the assistant. Uzun et al. [33] also argued that students with lower SRL competence engage with analytics feedback less than their high SRL peers, resulting in the need to cater for the diverse levels of competence among students. This study supported their findings and extended them by exploring how students with varying SRL levels differed in their queries to the GenAI assistant and in which aspects they required assistance to engage and act on feedback. Thus, it contributes to the understanding of how conversational analytics systems can be designed and used to address both the interpretive and affective needs of low SRL learners and evaluative and strategic practices of high SRL learners. Requests for feedback on artefacts, such as debate posts or written reflections, further revealed SRL level differences. Low SRL students sought reassurance about whether their outputs met expectations, compared to high SRL students who focused on more targeted improvements. Prior research has shown that one limitation of LADs is their focus on activity metrics rather than the quality of student work [2, 25]. Our findings confirmed this limitation and demonstrated that students at all SRL levels want richer, task-level feedback integrated into the dashboard and ready to bypass this constraint by asking the assistant for feedback on their artefacts. Beyond task-level feedback, students also used clarification queries as a proxy for trust-building. They frequently asked how their data was collected and processed, seeking reassurance that feedback was reflecting real experience before acting on it [39]. This negotiation of trust then evolved into justification queries, where students highlighted mismatches in their data, explained their behaviours, defended their decisions, or referenced their lived experience that the system failed to capture [15]. These findings reflected active sense-making and aligned with literature on dialogic feedback, where learners co-construct meaning and take greater ownership of their progress [10, 39]. Such dialogues demonstrated how conversational systems can promote deeper engagement with feedback and encourage students to critically evaluate their strategies [6, 14]. A responsive assistant should acknowledge and accommodate students' contributions and adapt responses accordingly (e.g., "*Thanks for clarifying! Let's adjust your study plan based on...*").

Drawing on the query types identified in RQ1, **RQ2** focused on how effectively the assistant handled these varied demands. Across interactions, the assistant's greatest strengths were in clarity and reliability, particularly in queries where students sought to understand or justify their feedback. Consistent with prior work [9], generative models proved effective at structuring information into coherent, accessible explanations, making them well-suited for clarifying analytics metrics and addressing misunderstandings in the early stages of sense-making. Especially, in justification and emotion-themed queries, this often translated into dialogue that helped students negotiate meaning and contextualise feedback within their lived experiences. However, the results also made clear that clarity alone was not enough for meaningful pedagogical support. In high-demand categories such as personalised insights and more content information, the assistant frequently offered generic recommendations rather than tailoring advice to individual performance data. This limitation mirrored the findings by Jin et al. [16] and Jung and Wise [17], who noted that feedback without specific, actionable guidance risks being perceived as redundant or unhelpful. Some students circumvented this issue by changing the prompt/ query from "*How can I improve in the debate activity?*" to "*Is there any specific areas of improvement **based on just my data?***" This simple yet powerful change in the query yielded different, satisfying outputs. However, this skill was not common among students, and it should not be solely the student's responsibility to elicit relevant responses. Similarly, moderate scores for context awareness suggested that while the assistant could interpret single data points such as metrics, it rarely synthesised information across multiple data points and information sources in the system, resulting in fragmented advice. To address these issues, two solutions stand out. First is strengthening the assistant side to synthesise data across sources so responses could be more tailored and actionable [30]. The second solution is to teach students how to query well-constructed, context-rich prompts, which can significantly enhance the quality of student-GenAI interactions in higher education [22, 23]. Given the scale and black-box nature of large language models, even small changes in phrasing, as exemplified above, can shift the model's attention to a different context and change the output [21]. Building this prompt literacy could improve personalisation and the effectiveness of the assistant's responses. A further challenge emerged in handling emotionally charged queries. While the assistant generally maintained a supportive tone, receiving high motivation and engagement ratings, it was less capable of responding to affective cues. The study of Jansen et al. [12] has shown that students' emotional reactions to feedback strongly shape their understanding and motivation. Also, our findings extended this concern to the GenAI implemented feedback systems: the difficulty of automating socio-emotional feedback, despite its critical role in sustaining engagement and supporting emotional regulation [24]. Without the capacity to detect and adapt to the emotional states of students, the assistant may overlook the opportunities to provide timely, empathic support that reinforces engagement and self-regulation.

Regarding **RQ3**, the students' reflections presented a nuanced picture of the GenAI assistant's role in their analytics feedback interpretation, revealing its potential to scaffold SRL competencies while exposing limitations that hindered its full impact. Many students valued the personalised, data-driven recommendations, such as structured study routines or targeted resources, that linked directly to their analytics patterns, supporting reflection and adaptive strategy use [24, 32, 33]. This aligns with recent evidence that GenAI agents can strengthen learners' comprehension of complex visual analytics [37]. At the same time, some features of the GenAI assistant limited its effectiveness, and students' critiques exposed deeper tensions between analytics design, student agency, and trust [8, 19]. First, the analysis of authentic conversations uncovered key factors driving the observed discrepancies. The assistant often acted as a mediator between students' study realities and the data presented in the LAD. For example, questions about offline study habits, like "*How accurate are the measurements if I open the materials once and download them?*", exposed gaps in the system's ability to represent actual effort. When such gaps remain unaddressed, they can undermine trust in both the perceived usefulness of GenAI-implemented feedback and the accuracy of the underlying data, with platform-centric metrics risking alienation when they fail to reflect students' learning realities. In contrast, the assistant's acknowledgement and accurate responses to these blind spots helped validate diverse learning behaviours and demonstrated how conversational agents can foster transparency and trust in LADs. As Jin et al. [13] similarly reported that discrepancies between students' expectations and actual engagement with the GenAI feedback system can reduce sustained interaction over time. Therefore, LADs may benefit from incorporating more comprehensive forms of evidence, such as self-reports, while retaining a GenAI assistant as a mediator to bridge gaps between data, expectations, and learning practices. Second, students' frustration with generic or repetitive responses pointed to a lack of adaptivity in the assistant's system design. While its positivity was motivational, some learners, especially those with high SRL competence and strong awareness of their learning behaviours, perceived responses as overly cautious and insufficiently challenging. This experience aligns with how novice and expert students differ in their approach to learning: feedback that feels safe may support novice students, but limit growth for advanced learners [41]. GenAI assistants should calibrate depth and tone of responses dynamically to match students' expertise and self-awareness. Finally, students approached these challenges as opportunities to refine the LAD and GenAI assistant. They envisioned an assistant capable of moving beyond static interpretation toward proactive and reflective questioning to their queries, integrated with course activities and able to adapt to prior learning histories and background [15, 37]. This vision reflects dialogic feedback principles [6] and aligns with Banihashem et al.'s [1] framework for hybrid intelligent feedback, which advocates AI systems that initiate metacognitive conversations, support goal setting, and connect analytics feedback to broader learning trajectories. Technical enhancements, such as faster processing, conversation history, and longitudinal tracking, were seen as essential for achieving this shift from feedback provider to active learning partner.

## 5 LIMITATIONS AND FUTURE WORK

While this study provides valuable results on the role of a GenAI assistant in supporting learning analytics feedback provision, it has certain limitations, which bear opportunities for future research. First, although the assistant was available throughout the semester, only 22 of the 34 students engaged with it. The conversational dataset was therefore relatively small and drawn from a single course. This constrains the generalisability of findings and positions them as exploratory rather than confirmatory. Patterns observed here may differ in courses with larger enrolments, alternative pedagogical designs, or different cultural contexts. Future work should replicate and extend the study across diverse cohorts, disciplines, and institutional settings to examine the stability of these findings, as well as explore why a subset of students did not engage with the assistant. Second, the study context involved both a specific LAD and a single GenAI configuration, meaning that the dashboard's available data, visualisation style, and the assistant's prompt design and capabilities collectively shaped the nature of interactions between the student and the assistant. The application's technical limitations, particularly the absence of conversational memory, further constrained responsiveness, adaptability, and longitudinal support. Without the ability to reference prior knowledge exchanges, the assistant was unable to sustain coherent, goal-oriented dialogue over time. These constraints pointed to the potential value of alternative AI system configurations. For example, richer analytics datasets, multimodal visualisations, fine-tuned AI models, or memory-enabled systems could produce distinct engagement patterns and perceptions [40]. Improving these capabilities would align with Banihashem et al.'s [1] emphasis on iterative refinement and dynamic adaptability in hybrid intelligent feedback systems. Features such as multi-turn scaffolding mechanisms [37] and persistent dialogue histories could transform isolated exchanges of information into a sustained, adaptive learning partnership. Third, this study relied on post-hoc analysis of chat transcripts and reflective writings. While effective for identifying themes and patterns, this approach did not allow for real-time intervention or adaptive support. Integrating real-time automatic conversational analysis into the LAD could enable automated detection of patterns such as repeated requests, affective cues like frustration, or low-quality responses [1]. Prior work in learning analytics and educational data mining has demonstrated the feasibility of real-time detectors for student affect, disengagement, and help-seeking behaviour, enabling timely pedagogical responses [10, 28, 32]. Finally, the study did not directly measure changes in academic performance, engagement, or other behavioural outcomes [33]. Without such measures, the impact of GenAI-assisted feedback remained inferred from interaction patterns and self-reported perceptions. Incorporating assessments or learning traces in future work would strengthen the claims about its educational value.

## 6 CONCLUSION

In this study, postgraduate students' authentic interactions with a custom-built GenAI assistant embedded within LAD were investigated in a semester-long course context. It was aimed to shed light on how students with different levels of SRL competence interpret, challenge, and act upon analytics feedback when given the opportunity to converse with it. The findings showed that in both levels of SRL, students did not treat the GenAI assistant as static output but as a springboard for dialogue, seeking clarification, offering justification, and testing the feedback system's reliability. Low SRL students were primarily seeking clarification and reassurance, while high SRL students instead focused on technical queries and requested personalised strategies. The assistant was generally effective in delivering clear and trustworthy responses, but its limitations in personalisation, contextual synthesis, and emotional responsiveness constrained its impact on deeper learning regulation. The findings of this study suggest that clarity and coherence are necessary but not sufficient qualities for GenAI scaffolded feedback. For such systems to further support engagement with feedback in line with students' SRL competence, they should evolve toward greater responsiveness, adaptivity, and relational continuity both technically and pedagogically. Ultimately, students showed a strong awareness of both the potential and the limits of generative feedback, offering concrete, forward-looking suggestions that pointed the way to more meaningful GenAI partnerships in education.

**ACKNOWLEDGMENTS**

We would like to thank DUTE 2024/2025 students for granting permission to collect data for this study. Yildiz Uzun was funded by the Republic of Türkiye Ministry of National Education for her postgraduate research.

**REFERENCES**
[1] Banihashem, S.K. et al. 2025. Pedagogical framework for hybrid intelligent feedback. *Innovations in Education and Teaching International*. 0, 0 (2025), 1–17. https://doi.org/10.1080/14703297.2025.2499174.
[2] Bodily, R. et al. 2018. The design, development, and implementation of student-facing learning analytics dashboards. *Journal of Computing in Higher Education*. 30, 3 (Dec. 2018), 572–598. https://doi.org/10.1007/s12528-018-9186-0.
[3] Bowman, D. et al. 2021. The Mathematical Foundations of Epistemic Network Analysis. *Communications in Computer and Information Science*. 1312, (Jan. 2021), 91–105. https://doi.org/10.1007/978-3-030-67788-6_7.
[4] Braun, V. and Clarke, V. 2022. *Thematic analysis: a practical guide*. SAGE.
[5] Braun, V. and Clarke, V. 2006. Using thematic analysis in psychology. *Qualitative Research in Psychology*. 3, 2 (Jan. 2006), 77–101. https://doi.org/10.1191/1478088706qp063oa.
[6] Carless, D. and Boud, D. 2018. The development of student feedback literacy: enabling uptake of feedback. *Assessment & Evaluation in Higher Education*. 43, 8 (Nov. 2018), 1315–1325. https://doi.org/10.1080/02602938.2018.1463354.
[7] Carless, D. and Winstone, N. 2023. Teacher feedback literacy and its interplay with student feedback literacy. *Teaching in Higher Education*. 28, 1 (Jan. 2023), 150–163. https://doi.org/10.1080/13562517.2020.1782372.
[8] Cukurova, M. 2024. The interplay of learning, analytics and artificial intelligence in education: A vision for hybrid intelligence. *British Journal of Educational Technology*. (Aug. 2024), bjet.13514. https://doi.org/10.1111/bjet.13514.
[9] Dai, W. et al. 2024. Assessing the proficiency of large language models in automatic feedback generation: An evaluation study. *Computers and Education: Artificial Intelligence*. 7, (Dec. 2024), 100299. https://doi.org/10.1016/j.caeai.2024.100299.

[10] Darvishi, A. et al. 2024. Impact of AI assistance on student agency. *Computers & Education*. 210, (Mar. 2024), 104967. https://doi.org/10.1016/j.compedu.2023.104967.

[11] Hattie, J. and Timperley, H. 2007. The Power of Feedback. *Review of Educational Research*. 77, 1 (Mar. 2007), 81–112. https://doi.org/10.3102/003465430298487.

[12] Jansen, T. et al. 2025. Constructive feedback can function as a reward: Students' emotional profiles in reaction to feedback perception mediate associations with task interest. *Learning and Instruction*. 95, (Feb. 2025), 102030. https://doi.org/10.1016/j.learninstruc.2024.102030.

[13] Jin, F. et al. 2025. Students' Perceptions of GenAI-powered Learning Analytics in the Feedback Process: : A Feedback Literacy Perspective. *Journal of Learning Analytics*. (Mar. 2025), 1–17. https://doi.org/10.18608/jla.2025.8609.

[14] Jin, H. et al. 2022. Towards Supporting Dialogic Feedback Processes Using Learning Analytics: the Educators' Views on Effective Feedback. *ASCILITE Publications*. (Nov. 2022), e22054. https://doi.org/10.14742/apubs.2022.54.

[15] Jin, Y. et al. 2025. Chatting with a Learning Analytics Dashboard: The Role of Generative AI Literacy on Learner Interaction with Conventional and Scaffolding Chatbots. *Proceedings of the 15th International Learning Analytics and Knowledge Conference* (Dublin Ireland, Mar. 2025), 579–590.

[16] Jin, Y. et al. 2025. Generative AI in higher education: A global perspective of institutional adoption policies and guidelines. *Computers and Education: Artificial Intelligence*. 8, (June 2025), 100348. https://doi.org/10.1016/j.caeai.2024.100348.

[17] Jung, Y. and Wise, A.F. 2024. Probing Actionability in Learning Analytics: The Role of Routines, Timing, and Pathways. *Proceedings of the 14th Learning Analytics and Knowledge Conference* (Kyoto Japan, Mar. 2024), 871–877.

[18] Kaliisa, R. et al. 2024. Have Learning Analytics Dashboards Lived Up to the Hype? A Systematic Review of Impact on Students' Achievement, Motivation, Participation and Attitude. *Proceedings of the 14th Learning Analytics and Knowledge Conference* (Kyoto Japan, Mar. 2024), 295–304.

[19] Khalil, M. and Prinsloo, P. 2025. The lack of generalisability in learning analytics research: why, how does it matter, and where to? *Proceedings of the 15th International Learning Analytics and Knowledge Conference* (Dublin Ireland, Mar. 2025), 170–180.

[20] Khosravi, H. et al. 2025. Generative AI and Learning Analytics: Pushing Boundaries, Preserving Principles. *Journal of Learning Analytics*. 12, 1 (Mar. 2025), 1–11. https://doi.org/10.18608/jla.2025.8961.

[21] Knoth, N. et al. 2024. AI literacy and its implications for prompt engineering strategies. *Computers and Education: Artificial Intelligence*. 6, (June 2024), 100225. https://doi.org/10.1016/j.caeai.2024.100225.

[22] Koyuturk, C. et al. 2025. Understanding Learner-LLM Chatbot Interactions and the Impact of Prompting Guidelines. *Artificial Intelligence in Education* (Cham, 2025), 364–377.

[23] Lee, D. and Palmer, E. 2025. Prompt engineering in higher education: a systematic review to help inform curricula. *International Journal of Educational Technology in Higher Education*. 22, 1 (Feb. 2025), 7. https://doi.org/10.1186/s41239-025-00503-7.

[24] Lim, L.-A. et al. 2021. Students' perceptions of, and emotional responses to, personalised learning analytics-based feedback: an exploratory study of four courses. *Assessment & Evaluation in Higher Education*. 46, 3 (Apr. 2021), 339–359. https://doi.org/10.1080/02602938.2020.1782831.

[25] Matcha, W. et al. 2020. A Systematic Review of Empirical Studies on Learning Analytics Dashboards: A Self-Regulated Learning Perspective. *IEEE Transactions on Learning Technologies*. 13, 2 (Apr. 2020), 226–245. https://doi.org/10.1109/TLT.2019.2916802.

[26] McHugh, M. 2012. Interrater reliability: The kappa statistic. *Biochemia medica : časopis Hrvatskoga društva medicinskih biokemičara / HDMB*. 22, (Oct. 2012), 276–82. https://doi.org/10.11613/BM.2012.031.

[27] Molenaar, I. et al. 2023. Measuring self-regulated learning and the role of AI: Five years of research using multimodal multichannel data. *Computers in Human Behavior*. 139, (Feb. 2023), 107540. https://doi.org/10.1016/j.chb.2022.107540.

[28] Ramaswami, G. et al. 2023. Use of Predictive Analytics within Learning Analytics Dashboards: A Review of Case Studies. *Technology, Knowledge and Learning*. 28, 3 (Sept. 2023), 959–980. https://doi.org/10.1007/s10758-022-09613-x.

[29] Shaffer, D.W. et al. 2016. A Tutorial on Epistemic Network Analysis: Analyzing the Structure of Connections in Cognitive, Social, and Interaction Data. *Journal of Learning Analytics*. 3, 3 (Dec. 2016), 9–45. https://doi.org/10.18608/jla.2016.33.3.

[30] Stadler, M. et al. 2024. Cognitive ease at a cost: LLMs reduce mental effort but compromise depth in student scientific inquiry. *Computers in Human Behavior*. 160, (Nov. 2024), 108386. https://doi.org/10.1016/j.chb.2024.108386.
[31] Suraworachet, W. et al. 2021. Examining the Relationship Between Reflective Writing Behaviour and Self-regulated Learning Competence: A Time-Series Analysis. *Technology-Enhanced Learning for a Free, Safe, and Sustainable World*. T. De Laet et al., eds. Springer International Publishing. 163–177.
[32] Susnjak, T. et al. 2022. Learning analytics dashboard: a tool for providing actionable insights to learners. *International Journal of Educational Technology in Higher Education*. 19, 1 (Dec. 2022), 12. https://doi.org/10.1186/s41239-021-00313-7.
[33] Uzun, Y. et al. 2025. Engagement with analytics feedback and its relationship to self-regulated learning competence and course performance. *International Journal of Educational Technology in Higher Education*. 22, 1 (Mar. 2025), 17. https://doi.org/10.1186/s41239-025-00515-3.
[34] Verbert, K. et al. 2013. Learning Analytics Dashboard Applications. *American Behavioral Scientist*. 57, 10 (Oct. 2013), 1500–1509. https://doi.org/10.1177/0002764213479363.
[35] Verbert, K. et al. 2020. Learning analytics dashboards: the past, the present and the future. *Proceedings of the Tenth International Conference on Learning Analytics & Knowledge* (Frankfurt Germany, Mar. 2020), 35–40.
[36] Winne, P.H. and Hadwin, A.E. 1998. Studying as Self-Regulated Learning. *Metacognition in Educational Theory and Practice*. Erlbaum. 277–304.
[37] Yan, L. et al. 2025. The effects of generative AI agents and scaffolding on enhancing students' comprehension of visual learning analytics. *Computers & Education*. 234, (Sept. 2025), 105322. https://doi.org/10.1016/j.compedu.2025.105322.
[38] Yan, L. et al. 2024. VizChat: Enhancing Learning Analytics Dashboards with Contextualised Explanations Using Multimodal Generative AI Chatbots. *Artificial Intelligence in Education*. A.M. Olney et al., eds. Springer Nature Switzerland. 180–193.
[39] Yang, M. and Carless, D. 2013. The feedback triangle and the enhancement of dialogic feedback processes. *Teaching in Higher Education*. 18, 3 (Apr. 2013), 285–297. https://doi.org/10.1080/13562517.2012.719154.
[40] Zhang, X. et al. 2024. A Systematic Literature Review of Empirical Research on Applying Generative Artificial Intelligence in Education. *Frontiers of Digital Education*. 1, 3 (Sept. 2024), 223–245. https://doi.org/10.1007/s44366-024-0028-5.
[41] Zimmerman, B.J. 2000. Attaining Self-Regulation. *Handbook of Self-Regulation*. Elsevier. 13–39.